\documentclass[letterpaper, 10 pt, journal, twoside]{IEEEtran}

\IEEEoverridecommandlockouts                              % This command is only needed if
\let\labelindent\relax

\usepackage{moreverb,url}
\usepackage{pgfplots}
\usepackage{pgfplotstable}
\usepgfplotslibrary{groupplots}
\usepackage{amsmath}
\usepackage{amsfonts}
\usepackage{amsthm}
\usepackage{import}
\usepackage{siunitx}
\usepackage{cleveref}
\usepackage{algorithm}
\usepackage{array}
\usepackage[noend]{algpseudocode}
\usepackage{enumitem}
\setitemize{noitemsep,topsep=0pt,parsep=0pt,partopsep=0pt}
\crefformat{figure}{Fig.~#2#1#3}
\Crefformat{figure}{Figure~#2#1#3}
\crefformat{equation}{(#2#1#3)}
\crefformat{section}{Section~#2#1#3}
\crefformat{table}{Table~#2#1#3}
\crefformat{algorithm}{Algorithm~#2#1#3}
\crefformat{appendix}{Appendix~#2#1#3}

\DeclareRobustCommand{\vec}[1]{
	\ifthenelse{\equal{#1}{\omega} \OR \equal{#1}{\varphi} \OR \equal{#1}{\alpha} \OR \equal{#1}{\beta} \OR \equal{#1}{\chi} \OR \equal{#1}{\delta} \OR \equal{#1}{\varepsilon} \OR \equal{#1}{\phi} \OR \equal{#1}{\epsilon} \OR \equal{#1}{\gamma} \OR \equal{#1}{\eta} \OR \equal{#1}{\iota} \OR \equal{#1}{\kappa} \OR \equal{#1}{\lambda} \OR \equal{#1}{\mu} \OR \equal{#1}{\nu} \OR \equal{#1}{\pi} \OR \equal{#1}{\theta} \OR \equal{#1}{\vartheta} \OR \equal{#1}{\rho} \OR \equal{#1}{\sigma} \OR \equal{#1}{\varsigma} \OR \equal{#1}{\tau} \OR \equal{#1}{\upsilon} \OR \equal{#1}{\xi} \OR \equal{#1}{\psi} \OR \equal{#1}{\zeta}}{
		\boldsymbol{#1}
	}{
		\mathbf{#1}
	}
}

\DeclareMathOperator*{\argmin}
{arg\,min}

\usetikzlibrary {3d}
\usetikzlibrary{decorations.markings}
\usepgfplotslibrary{fillbetween}
\usepgfplotslibrary{patchplots}
\usetikzlibrary{external}
\usetikzlibrary {arrows.meta}
\pgfplotsset{every axis/.append style={
			font=\footnotesize,
			line join=round,
			legend style={/tikz/every even column/.append style={column sep=0.2cm}}
		}}

\DeclareRobustCommand{\qs}[1]{\textsinglequote s}

\renewcommand{\dddot}[1]{%
	{\mathop{\kern\z@#1}\limits^{\makebox[0pt][c]{\vbox to-1.4\ex@{\kern-\tw@\ex@
						\hbox{\normalfont...}\vss}}}}}
\newcommand{\dddotsuper}[1]{%
	{\mathop{\kern\z@#1}\limits^{\makebox[0pt][c]{\vbox to-1.6\ex@{\kern-\tw@\ex@
						\hbox{\scriptsize...}\vss}}}}}
\newcommand{\dddotalt}[1]{%
	{\mathop{\kern\z@#1}\limits^{\makebox[0pt][c]{\vbox to-2.2\ex@{\kern-\tw@\ex@
						\hbox{\normalfont\scaleddot\kern-0.5pt\scaleddot\kern-0.5pt\scaleddot}\vss}}}}}
\makeatother

\newtheorem*{remark}{Remark}
\setitemize{noitemsep,topsep=0pt,parsep=0pt,partopsep=0pt}

\title{
BoundPlanner: A convex-set-based approach \\ to bounded manipulator trajectory planning
}

\author{Thies Oelerich$^{1}$, Christian Hartl-Nesic$^{1}$, Florian Beck$^{1}$, Andreas Kugi$^{1, 2}$% <-this % stops a space
\thanks{Manuscript received: December, 13, 2024; Revised March, 6, 2025; Accepted March, 27, 2025.}%
\thanks{This paper was recommended for publication by Editor Hyungpil Moon upon evaluation of the Associate Editor and Reviewers comments}
\thanks{*The authors acknowledge TU Wien Bibliothek for financial support through its Open Access Funding Program..}% <-this % stops a space
\thanks{$^{1}$All authors are with the Automation and Control Institute (ACIN), TU
    Wien, Vienna, Austria,
    {\tt\small \{oelerich, hartl, beck, kugi\}@acin.tuwien.ac.at}}%
\thanks{$^{2}$Andreas Kugi is with the
    AIT Austrian Institute of Technology GmbH, Vienna, Austria,
{\tt\small andreas.kugi@ait.ac.at}}%
}

\newcommand{\review}[1] {\textcolor{black}{#1}}

\begin{document}

\markboth{IEEE Robotics and Automation Letters. Preprint Version. Accepted April, 2025}
{Oelerich \MakeLowercase{\textit{et al.}}: BoundPlanner}

\maketitle

% This forces figure 1 to be on the right column
\global\csname @topnum\endcsname 0
\global\csname @botnum\endcsname 0

%%%%%%%%%%%%%%%%%%%%%%%%%%%%%%%%%%%%%%%%%%%%%%%%%%%%%%%%%%%%%%%%%%%%%%%%%%%%%%%%
\begin{abstract}
	Online trajectory planning enables robot manipulators to react quickly to changing environments or tasks. Many robot trajectory planners exist for known environments but are often too slow for online computations. Current methods in online trajectory planning do not find suitable trajectories in challenging scenarios that respect the limits of the robot and account for collisions. This work proposes a trajectory planning framework consisting of the novel Cartesian path planner based on convex sets, called BoundPlanner, and the online trajectory planner BoundMPC~\cite{oelerichBoundMPCCartesianTrajectory2024}. BoundPlanner explores and maps the collision-free space using convex sets to compute a reference path with bounds. BoundMPC is extended in this work to handle convex sets for path deviations, which allows the robot to optimally follow the path within the bounds while accounting for the robot\textquotesingle s kinematics. Collisions of the robot\textquotesingle s kinematic chain are considered by a novel convex-set-based collision avoidance formulation independent on the number of obstacles.
	Simulations and experiments with a 7-DoF manipulator show the performance of the proposed planner compared to state-of-the-art methods.
	The source code is available at~\url{github.com/TU-Wien-ACIN-CDS/BoundPlanner} and videos of the experiments can be found at~\url{www.acin.tuwien.ac.at/42d4}.
	% TODO videos and code publish
\end{abstract}


%%%%%%%%%%%%%%%%%%%%%%%%%%%%%%%%%%%%%%%%%%%%%%%%%%%%%%%%%%%%%%%%%%%%%%%%%%%%%%%%
\section{Introduction}

\begin{figure}[t]
	\centering
	% \addtolength\abovecaptionskip{-5pt}
	\def\svgwidth{0.9\linewidth}
	% \import{inkscape}{schematic.pdf_tex}
	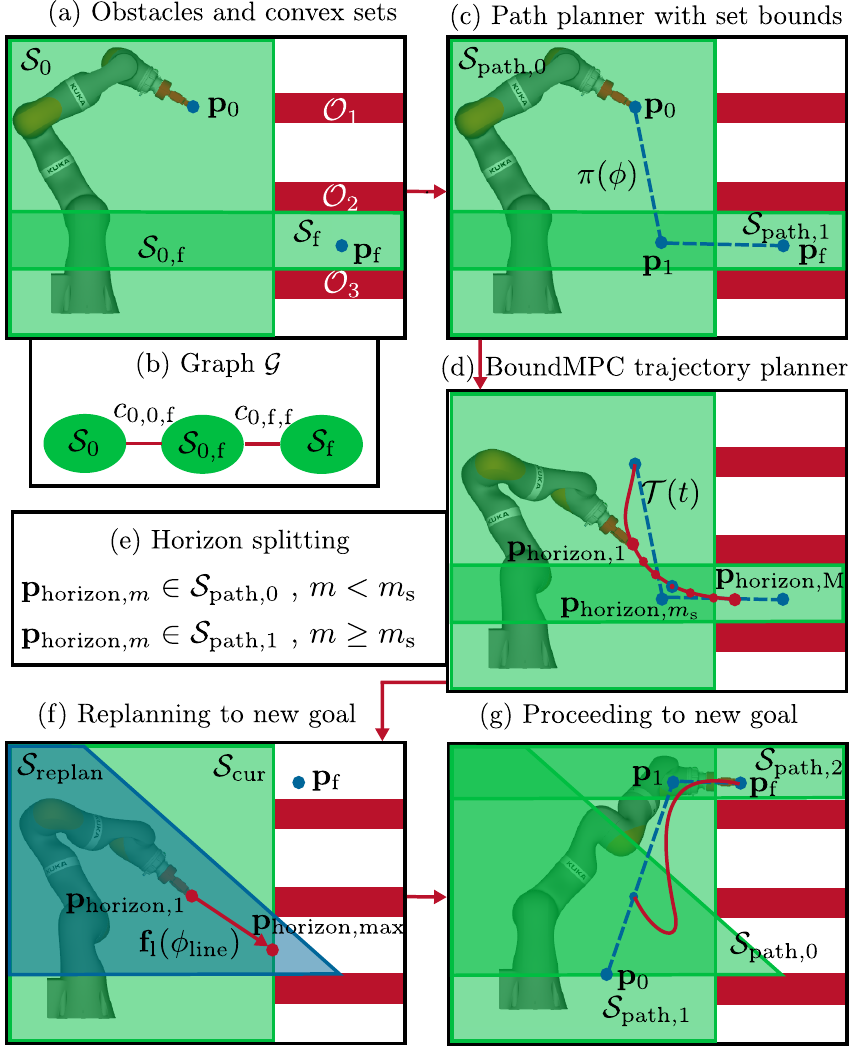
	\caption{Schematic of the path planning using \textit{BoundPlanner} to plan a Cartesian path $\pi(\phi)$ from the start point $\vec{p}_{0}$ to the end point $\vec{p}_{\mathrm{f}}$: (a) The graph $\mathcal{G}$ is built using the convex sets $\mathcal{S}_{0}$ and $\mathcal{S}_\mathrm{f}$ around $\vec{p}_{0}$, $\vec{p}_{\mathrm{f}}$, further visualized in (b). (c) A sequence of convex sets $\mathcal{S}_{\mathrm{path}, 0}, \mathcal{S}_{\mathrm{path}, 1}$ is found using $\mathcal{G}$, connecting the starting set $\mathcal{S}_{0}$ with the ending set $\mathcal{S}_{\mathrm{f}}$. A path $\pi(\phi)$ is constructed inside these sets connecting $\vec{p}_{0}$ with $\vec{p}_{\mathrm{f}}$. (d) BoundMPC~\cite{oelerichBoundMPCCartesianTrajectory2024} uses the path $\pi(\phi)$ of \textit{BoundPlanner} to find a suitable joint trajectory $\mathcal{T}(t)$ such that the end-effector stays within the convex sets using horizon splitting (e) at index $m_{\mathrm{s}}$ of the finite horizon $\vec{p}_{\mathrm{horizon}, 1}, \ldots, \vec{p}_{\mathrm{horizon, M}}$, thus avoiding the obstacles $\mathcal{O}_{\mathrm{1}}, \mathcal{O}_{\mathrm{2}}, \mathcal{O}_{\mathrm{3}}$. (f) A new goal $\vec{p}_\mathrm{f}$ requires a replanning using the replanning set $\mathcal{S}_\mathrm{replan}$ constructed based on the current set $\mathcal{S}_\mathrm{cur}$ and the line $\vec{f}_{\mathrm{l}}(\phi_{\mathrm{line}})$, which is based on the last position $\vec{p}_{\mathrm{horizon, max}}$ in the planning horizon that is within $\mathcal{S}_{\mathrm{cur}}$. With $\mathcal{S}_\mathrm{replan}$ as the new initial set $\mathcal{S}_0$, a new trajectory is planned in (g) to $\vec{p}_\mathrm{f}$ as in (a)-(d).}
	\label{fig:planner_scheme}
\end{figure}

\IEEEPARstart{M}{odern} robots are equipped with numerous sensors enabling them to act autonomously in unknown situations and solve tasks requiring flexibility like household tasks~\cite{oelerichLanguageguidedManipulatorMotion2024}. For example, tidying up a room is challenging due to the unstructured environment in which the robot has to act. It is infeasible to provide a complete description of such an environment as it is prone to changes. Obstacles like chairs, tables, and humans may change locations at any time. The robot needs to move fluently and quickly, which presents a major challenge for trajectory planning.
Planning robot motions is difficult due to kinematic, dynamic and collision-avoidance constraints making offline planners~\cite{lavallePlanningAlgorithms2006,marcucciMotionPlanningObstacles2023} a common choice.
However, it is important to plan robotic motions online and react quickly in order to address the challenges of unstructured and dynamic environments.
Popular approaches are online optimization-based trajectory planning with a receding horizon~\cite{oelerichBoundMPCCartesianTrajectory2024,oelerichLanguageguidedManipulatorMotion2024, beckModelPredictiveTrajectory2024, oelerichModelPredictiveTrajectory2024} or sampling-based trajectory planning~\cite{bhardwajSTORMIntegratedFramework2022,jankowskiVPSTOPointbasedStochastic2023}.

As using a finite planning horizon limits the optimality of the resulting trajectory,
the trajectory planning problem may be separated into two subproblems by first computing a reference path as a geometrical curve in space and then finding a trajectory that follows this reference path~\cite{oelerichBoundMPCCartesianTrajectory2024, verschuerenTimeoptimalMotionPlanning2016}. The reference path, computed by, e.g.,~\cite{perssonSamplingbasedAlgorithmRobot2014, oelerichLanguageguidedManipulatorMotion2024}, provides global guidance for the subsequent path-following controller~\cite{oelerichBoundMPCCartesianTrajectory2024, vanduijkerenPathfollowingNMPCSeriallink2016, hartl-nesicSurfacebasedPathFollowing2021}. If the reference path considers the robot\textquotesingle s kinematics and collisions, the path may be followed exactly, but such a path is difficult to compute. Therefore, the works~\cite{oelerichBoundMPCCartesianTrajectory2024, romeroModelPredictiveContouring2022, arrizabalagaTimeOptimalTunnelFollowingQuadrotors2022} propose to allow path deviations. This idea is also used in~\cite{oelerichBoundMPCCartesianTrajectory2024} and~\cite{oelerichLanguageguidedManipulatorMotion2024} to follow Cartesian paths under consideration of the robot\textquotesingle s kinematics. Bounds around the path ensure collision freedom while allowing the trajectory planner to deviate from the path to account for kinematic constraints. The work in~\cite{oelerichBoundMPCCartesianTrajectory2024, oelerichLanguageguidedManipulatorMotion2024} shows that this is advantageous when executing the robot\textquotesingle s motion compared to following a Cartesian path exactly.

Instead of describing the obstacles directly~\cite{beckModelPredictiveTrajectory2024}, a parametrization of the collision-free space can be used for trajectory planning, e.g., differentiable bounds along a pre-computed path~\cite{arrizabalagaDifferentiableCollisionFreeParametric2024}, grids~\cite{gaoOnlineSafeTrajectory2018}, or a graph of convex sets~\cite{marcucciMotionPlanningObstacles2023, suhFastSafeMotion2020, liCollisionFreeTrajectoryOptimization2024, wangFastIterativeRegion2024, liuPlanningDynamicallyFeasible2017}, see also~\cite{kulathungaSurveyMotionPlanning2023} for a related survey. Convex sets are particularly simple to compute in the Cartesian position space and scale well for environments with many obstacles, e.g., point clouds~\cite{wangFastIterativeRegion2024}. Graphs of convex sets are, e.g., computed using set overlaps~\cite{liCollisionFreeTrajectoryOptimization2024} or from a given (initial) path or trajectory~\cite{liuPlanningDynamicallyFeasible2017}. This approach was used for navigation of mobile robots~\cite{liCollisionFreeTrajectoryOptimization2024, kurtzTemporalLogicMotion2023} and drones~\cite{marcucciMotionPlanningObstacles2023, liuPlanningDynamicallyFeasible2017}, which are often modeled with simple configuration spaces, e.g., as point robots~\cite{suhFastSafeMotion2020, kurtzTemporalLogicMotion2023}. A smooth trajectory or path is then computed by planning parametrized curves through the graph of convex sets~\cite{marcucciMotionPlanningObstacles2023, liuPlanningDynamicallyFeasible2017, kurtzTemporalLogicMotion2023, gaoOnlineSafeTrajectory2018}.
This approach is not suitable for complicated configuration spaces. Nevertheless, it has also been used to plan trajectories for serial manipulators~\cite{marcucciMotionPlanningObstacles2023} and humanoids~\cite{kurtzTemporalLogicMotion2023}.
Planning in Cartesian space neglects the robot\textquotesingle s kinematics. Hence, directly planning end-effector motions often yields trajectories that are generally not executable without considering the kinematic limitations.
To overcome this,~\cite{marcucciMotionPlanningObstacles2023} uses convex sets in the joint space. These are computationally expensive since obstacles in the task space are hard to transfer to the joint space.
This prohibits online planning.
Thus, replanning the path/trajectory online is limited to Cartesian sets and is demonstrated in~\cite{liuPlanningDynamicallyFeasible2017}. Other methods only replan within a given graph of convex sets~\cite{liuCDRTRRTRealtimeRapidly2025} or require a given trajectory~\cite{wangFastIterativeRegion2024}.

An often used approach in online planning are potential functions~\cite{jankowskiVPSTOPointbasedStochastic2023,beckModelPredictiveTrajectory2024,bhardwajSTORMIntegratedFramework2022}. Such a formulation generates costs in a cost function when the robot is close to a collision, but it does not guarantee collision avoidance and can lead to issues with, e.g., thin obstacles. Furthermore, the resulting cost terms depend on the number of obstacles that scale poorly to environments with many obstacles, e.g., when using point clouds.

This work proposes a trajectory planner for robot manipulators, consisting of a novel convex-set-based Cartesian path planner called \emph{BoundPlanner} and the MPC-based trajectory planner \emph{BoundMPC}. The former performs fast reference path planning for the end-effector in Cartesian space and allows reactive online replanning using convex collision-free sets.
A graph of convex sets is computed first, followed by the path generation as in~\cite{liCollisionFreeTrajectoryOptimization2024}. This has the advantage that larger regions in space can be favored to account for the kinematics. However, it requires computing an orientation path for which an efficient formulation is developed.
BoundPlanner extends previous work~\cite{liCollisionFreeTrajectoryOptimization2024, kurtzTemporalLogicMotion2023, marcucciMotionPlanningObstacles2023, liuPlanningDynamicallyFeasible2017} by favoring large convex sets and focusing on fast replanning.
BoundMPC computes a joint trajectory online to follow the path using the Cartesian reference path while allowing path deviations within the convex sets to consider the robot\textquotesingle s kinematics. The original formulation~\cite{oelerichBoundMPCCartesianTrajectory2024} uses bounding polynomials which are replaced by convex sets in this work as they are easier to compute.
The proposed combination of BoundPlanner and BoundMPC leverages the fast computations of Cartesian convex sets to allow reactive online replanning while still ensuring the kinematic feasibility of the trajectories. It thus combines the advantages of convex sets in Cartesian space and control in the joint space. \review{Other path-following trajectory planners can be used with adaptations instead of BoundMPC, but as discussed above and in~\cite{oelerichBoundMPCCartesianTrajectory2024}, BoundMPC offers advantages over existing planners.}
While the proposed framework only plans a reference path for the end-effector, collision avoidance of the remaining links is also ensured based on a novel formulation using collision-free convex sets for key points on the manipulator, \review{extending the formulation in~\cite{spahnCoupledMobileManipulation2021} by considering the prediction horizon of BoundMPC}.
This formulation does not weight task success with collision avoidance and avoids passing through thin obstacles.

The main contributions of this work are
\begin{itemize}
	\item a fast Cartesian path planner called \emph{BoundPlanner}, which explores the collision-free space by computing Cartesian convex sets,
	\item an extension of BoundMPC~\cite{oelerichBoundMPCCartesianTrajectory2024} to exploit Cartesian convex sets as path deviation bounds using horizon splitting,
	\item a novel collision avoidance formulation for the entire robot manipulator, \review{which considers the prediction horizon and is independent of the number of obstacles}.
\end{itemize}

\section{BoundPlanner}%
\label{sec:bound_planner}

The proposed method, called BoundPlanner, is explained in detail in this section.
The objective of BoundPlanner is to compute a reference path $\pi(\phi)$ with bounds. The bounds are given as collision-free convex sets around the path. Given the initial pose of the robot\textquotesingle s
end-effector with position $\vec{p}_{\mathrm{0}}$ and rotation matrix
$\vec{R}_{\mathrm{0}}$ and the desired final pose defined by
$\vec{p}_{\mathrm{f}}$ and $\vec{R}_{\mathrm{f}}$, a graph $\mathcal{G}$ of collision-free convex sets is created and utilized to compute $\pi(\phi)$ and the corresponding bounds.
A schematic overview of the method is shown in~\cref{fig:planner_scheme}.
The graph $\mathcal{G}$ is used to plan the Cartesian reference path $\pi(\phi)$ for the
robot\textquotesingle s end-effector with convex sets $\mathcal{S}_{\mathrm{path}, i}$ as Cartesian bounds.
This path and the bounds are then used by the model predictive
control (MPC) framework BoundMPC~\cite{oelerichBoundMPCCartesianTrajectory2024} in~\cref{ssub:mpc_planning} to compute the trajectory $\mathcal{T}(t)$ online by following the created reference path within the given
bounds to reach the goal. The convex sets used for the bounds ensure the motion is restricted to
the collision-free space.

\subsection{Collision-free Convex Sets}%
\label{sub:convex_sets}

This section describes the procedure of finding collision-free convex sets in the Cartesian position space in a bounded domain $\mathcal{D} \subset \mathbb{R}^{3}$ with $K$ convex obstacles $\mathcal{O}_{k}$, $k = 1, \ldots, K$. In this work, a convex position set $\mathcal{S}$ is defined by a set of $S$ half spaces parametrized by a matrix $\vec{A} \in \mathbb{R}^{S\times 3}$ and a vector $\vec{b} \in \mathbb{R}^{S}$, i.e., $\mathcal{S} = \{ \vec{x} \in \mathcal{D} : \vec{A} \vec{x} \leq \vec{b} \}$.
The following subsections describe three ways to obtain such convex sets, i.e., two methods based on ellipsoid volume maximization, and one method using a given set of points.

\subsubsection{Ellipsoid Volume Maximization}
A convex set $\mathcal{S}$ defines a unique ellipsoid $\mathcal{E}$, specified by its shape matrix $\vec{C}_\mathcal{E}$ and midpoint $\vec{p}_{\mathcal{E}}$, inscribed into $\mathcal{S}$ with maximum volume, i.e., $\mathcal{E}$ is the maximum-volume inscribed ellipsoid (MVIE).
Computing $\mathcal{E}$ in $\mathcal{S}$ is a convex optimization problem that can be solved efficiently.
The works~\cite{deitsComputingLargeConvex2015} and~\cite{wangFastIterativeRegion2024} use an iterative procedure to find large collision-free convex sets using MVIE as follows. First, an initial collision-free ellipsoid is specified around a given seed point in the collision-free space. Second, an initial convex set is computed for this ellipsoid using MVIE. Third, an iterative procedure is employed in which the ellipsoid $\mathcal{E}$ and the convex set $\mathcal{S}$ are computed alternatingly. For more information, the reader is referred to~\cite{deitsComputingLargeConvex2015} and~\cite{wangFastIterativeRegion2024}.
The algorithm in~\cite{wangFastIterativeRegion2024} uses a second-order cone optimization which is considerably faster than the semidefinite programming approach~\cite{deitsComputingLargeConvex2015}; hence~\cite{wangFastIterativeRegion2024} is used in this work.
Two different variants of finding $\mathcal{S}$ are considered, namely
\begin{itemize}
	\item \textit{MVIE} optimizes all values of the ellipsoid $\mathcal{E}$, i.e., $\vec{C}_{\mathcal{E}}$ and $\vec{p}_{\mathcal{E}}$,
	\item \textit{MVIE-fixed-mid} optimizes only $\vec{C}_{\mathcal{E}}$ of the ellipsoid $\mathcal{E}$, while $\vec{p}_{\mathcal{E}}$ remains fixed.
\end{itemize}
The \textit{MVIE-fixed-mid} problem may be thought of as finding the largest convex set $\mathcal{S}$ around the point $\vec{p}_{\mathcal{E}}$, whereas the \textit{MVIE} problem does not need to contain $\vec{p}_{\mathcal{E}}$.

\subsubsection{Convex Hull Set}
The problems \textit{MVIE} and \textit{MVIE-fixed-mid} cannot guarantee that a certain set of points is contained in the convex set $\mathcal{S}$. Therefore, a third way is introduced, i.e., finding a convex set around a set $\mathcal{S}_{\mathrm{ch}}$ defined as the convex hull of the points $\vec{p}_{\mathrm{ch}, j}$, $j = 1, \ldots, J$.
In order to compute the set $\mathcal{S}$, the points on an obstacle $\mathcal{O}_{k}$, $k = 1, \ldots, K$, and the set $\mathcal{S}_{\mathrm{ch}}$ with the smallest distance are determined using
\begin{equation}
	\argmin_{\vec{p}_{\mathrm{ch}}^{*}\in \mathcal{S}_{\mathrm{ch}}, \vec{p}_{\mathcal{O}_{k}}^{*}\in \mathcal{O}_{k}} \lVert\vec{p}_{\mathcal{O}_{k}}^{*} - \vec{p}_{\mathrm{ch}}^{*}\rVert_{2}^{2}
	\label{eq:proj_conv_hull}
\end{equation}
The vector $\vec{a}_{k} = \vec{p}_{\mathcal{O}, k}^{*} - \vec{p}_{\mathrm{ch}}^{*}$ defines a half space to separate the obstacle $\mathcal{O}_{k}$ from the set $\mathcal{S}_{\mathrm{ch}}$. The corresponding half space distance is computed as $b_{k} = \vec{a}_{k}^{\mathrm{T}}\vec{p}_{\mathcal{O}, k}^{*}$ such that the half space touches the obstacle $\mathcal{O}_{k}$.
Together, $\vec{a}_{k}$  and $b_{k}$ constitute $\vec{A}$ and $\vec{b}$ and thus define $\mathcal{S}$. To improve the computational efficiency, the half spaces are computed starting with the closest obstacle. Obstacles further away are disregarded if the half space also separates the set $\mathcal{S}_{\mathrm{ch}}$ from such obstacles.
Generally, this procedure is computationally more efficient than the ellipsoid-based computations but does not optimize for the set volume.
This defines the third way of computing a collision-free set $\mathcal{S}$ in this work:
\begin{itemize}
	\item \textit{Set-convex-hull} compute $\mathcal{S}$ around the convex set $\mathcal{S}_{\mathrm{ch}}$.
\end{itemize}

\subsection{Graph of Collision-free Convex Sets}%
\label{sub:graph_convex_sets}

In this section, the graph $\mathcal{G}$ of convex sets and a path $\pi_{\mathcal{G}}$ through this graph are computed. The end-effector geometry is modeled as a convex hull defined by the set of points
$\vec{p}_{\mathrm{e}, l}(\phi)$, $l = 1, \ldots, L$. Hence, the path $\pi_{\mathcal{G}}$ connects the start set $\mathcal{S}_{\mathrm{0}}$, containing of the entire convex hull of the end-effector at its initial position $\vec{p}_{\mathrm{0}}$ with its initial orientation $\vec{R}_{0}$, with the final set $\mathcal{S}_{\mathrm{f}}$, containing the entire end-effector at its final position $\vec{p}_{\mathrm{f}}$ and orientation $\vec{R}_{\mathrm{f}}$.
To this end, the convex sets $\mathcal{S}_{\mathrm{0}}$ and $\mathcal{S}_{\mathrm{f}}$ are
computed using \textit{Set-convex-hull} around
the known end-effector hull in the known orientations $\vec{R}_{0}$ and $\vec{R}_{\mathrm{f}}$ at $\vec{p}_{0}$ and $\vec{p}_{\mathrm{f}}$, respectively.

\begin{figure}[t]
	\centering
	\def\svgwidth{0.55\linewidth}
	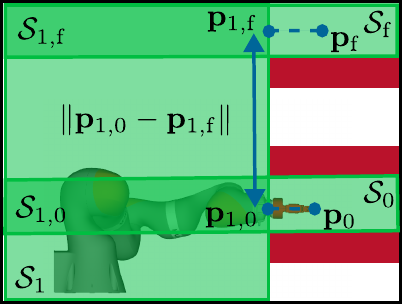
	\caption{Graph building procedure of BoundPlanner. The set $\mathcal{S}_{1}$ intersects with the starting set $\mathcal{S}_{0}$ and the final set $\mathcal{S}_{\mathrm{f}}$. In order to compute the costs $c_{\mathrm{0, 1, f}}$ from~\cref{eq:graph_cost} for the edge between the intersection sets $\mathcal{S}_{1, 0}$ and $\mathcal{S}_{\mathrm{1, f}}$, the projections $\vec{p}_{1, 0}$ and $\vec{p}_{\mathrm{1, f}}$ are computed using~\cref{eq:projection}.}
	\label{fig:graph_building}
\end{figure}

To find the structure of the collision-free space, multiple convex
sets are computed in the domain $\mathcal{D}$ and a graph $\mathcal{G}$ is built based on set intersections.
Two convex sets $\mathcal{S}_{a}$ and $\mathcal{S}_{b}$ intersect if their
intersection set $\mathcal{S}_{a, b}$ is not empty, i.e.,
$\mathcal{S}_{a, b} = \mathcal{S}_{a} \cap
	\mathcal{S}_{b} \neq \emptyset$.
In this work, the graph $\mathcal{G}$ describes the collision-free space with the intersection sets as its vertices and connecting sets as edges. An edge between two vertices indicates that the intersection sets (vertices) have a common set (edge) as part of the intersection.
The edge cost $c_{a, b, c}$
between the two intersection sets $\mathcal{S}_{a, b}$ and $\mathcal{S}_{b, c}$
\begin{equation}
	\label{eq:graph_cost}
	c_{a, b, c} =   c_{\mathrm{size}, b}\lVert \vec{p}_{a, b} - \vec{p}_{b, c} \rVert_{2} + c_{\mathrm{bias}}
\end{equation}
is proportional to the distance between the two intersection sets $\mathcal{S}_{a, b}$ and
$\mathcal{S}_{b, c}$,  approximated by the points $\vec{p}_{a, b} \in
	\mathcal{S}_{a, b}$ and $\vec{p}_{b, c} \in \mathcal{S}_{b, c}$, plus a
bias cost $c_{\mathrm{bias}} > 0$ to penalize set transitions. The size cost
$c_{\mathrm{size}, b}$ is chosen as
\begin{equation}
	c_{\mathrm{size}, b} = 1 + w_{\mathrm{size}}\tanh\left(\frac{1}{2} - \sqrt[3]{\det(\vec{C}_{\mathcal{E},b})}\right) \text{~,}
	\label{eq:size_cost}
\end{equation}
with $0 < w_{\mathrm{size}} \leq 1$ and the determinant of the MVIE in $\mathcal{S}_{b}$ as an approximation of the volume of $\mathcal{S}_{b}$. The formulation in~\cref{eq:size_cost} favors large sets, which allow larger path deviations when following the path. This way, BoundPlanner is specifically targeted for robot manipulators and their kinematic constraints.
Next assume that a new set $\mathcal{S}_a$ is added to the graph intersecting
with $\mathcal{S}_{b}$. Then the intersection set
$\mathcal{S}_{a, b}$ is assigned the point
\begin{equation}
	\vec{p}_{a, b} = \min_{\vec{p} \in \mathcal{S}_{a, b}} \lVert
	\vec{p} - \vec{p}_{b} \rVert_{2}^{2} \text{~,}
	\label{eq:projection}
\end{equation}
which is the projection of $\vec{p}_{b}$ onto the intersection set
$\mathcal{S}_{a,b}$.
Each intersection set
$\mathcal{S}_{a, b}$ is assigned a point $\vec{p}_{a, b}$ to
compute the edge cost~\cref{eq:graph_cost}.
The point $\vec{p}_{b}$ is equal to $\vec{p}_{0}$ or $\vec{p}_{\mathrm{f}}$ for the start and end sets, respectively, which is then iteratively projected when a new set is added to graph $\mathcal{G}$.
If neither the start set $\mathcal{S}_{0}$ nor the end set $\mathcal{S}_{\mathrm{f}}$ are intersected and no
prior projections are available, the two closest points in the intersection sets
are used.
The graph-building process is visualized in~\cref{fig:graph_building} for the example scenario in~\cref{fig:planner_scheme}.
Note that an intersection set is added to graph $\mathcal{G}$ in~\cref{fig:planner_scheme} only if the
end-effector fits inside the intersection set using the end-effector description introduced below in~\cref{sub:path_planning}.
% This is checked by sampling
% points within the intersection set and orientations on the geodesic between the
% start and end orientation and checking whether the end-effector is within the
% intersection set for the sampled pose.

\subsection{Exploring the Environment with Collision-free Convex Sets}%
\label{sub:explore_convex_sets}
Optimally, each point in the collision-free subspace $\mathcal{D}_{\mathrm{free}} \subset \mathcal{D}$ of the entire planning domain $\mathcal{D}$ is contained in at least one convex set in graph $\mathcal{G}$.
In order to explore $\mathcal{D}_{\mathrm{free}}$, the collision-free space is sampled and convex sets are computed around each sample point, which are further referred to as seed points. This sampling is performed until a path $\pi_{\mathcal{G}}$ connecting
$\mathcal{S}_{0}$ and $\mathcal{S}_{\mathrm{f}}$ is found in graph $\mathcal{G}$.
In this work, rejection sampling is used, and the
intersections of the computed convex sets are added to graph $\mathcal{G}$ using the procedure of the previous section.
The rejection sampling rejects seed points
within an obstacle $\mathcal{O}_{k}$ or within a previously
computed convex set. The sample will thus be in the unexplored space of $\mathcal{D}_{\mathrm{free}}$.
\begin{remark}
	This sampling approach is probabilistically complete, i.e., the collision-free subspace $\mathcal{D}_{\mathrm{free}} \subset \mathcal{D}$ will be covered by convex sets as the number of samples grows to infinity. Each seed point is contained in its respective convex set. Thus, if a solution exists, it will eventually be found.
\end{remark}
The path $\pi_{\mathcal{G}}$ is the shortest path of sets through the graph $\mathcal{G}$ using Dijkstra\textquotesingle s algorithm~\cite{dijkstraNoteTwoProblems1959}.
Once a valid path $\pi_{\mathcal{G}}$ is found, a refinement step is
performed. For this, new convex sets are
computed with the projection points $\vec{p}_{a, b}$ as seed points and are added
to $\mathcal{G}$. Successively, a new path $\pi_{\mathcal{G}}$ is computed through this graph. This procedure is
iteratively repeated until the path converges.
The optimal path $\pi_{\mathcal{G}}$ through the graph is finally obtained as a series of edges defined by the
sets $\mathcal{S}_{\mathrm{path}, i}$,
$i = 0, \ldots, N$, where
$\mathcal{S}_{\mathrm{path}, 0} = \mathcal{S}_0$ and
$\mathcal{S}_{\mathrm{path}, N} = \mathcal{S}_{\mathrm{f}}$.

An overview of the algorithm of BoundPlanner is given as pseudocode in~\cref{alg:bound_planner}.
Adding a new set $\mathcal{S}_{a}$ to the graph $\mathcal{G}$, including the computation of the intersections and costs, is represented by the function $\mathtt{add\_set\_to\_graph}(\mathcal{G}, \mathcal{S}_{a})$. It returns the $\mathtt{connected}$ flag, indicating if a connection between the start and the end set exists in graph $\mathcal{G}$.

\begin{algorithm}
	\small
	\caption{BoundPlanner}
	\label{alg:bound_planner}
	\begin{algorithmic}
		\Require Starting pose $\vec{p}_0$, $\vec{R}_0$ and final pose $\vec{p}_{\mathrm{f}}$, $\vec{R}_{\mathrm{f}}$, domain $\mathcal{D}$, obstacles $\mathcal{O}_{k}$
		\State $\mathcal{S}_0 \gets \text{\textit{Set-convex-hull} around } \vec{p}_0, \vec{R}_{0}$ \Comment{Set around start}
		\State $\mathcal{S}_{\mathrm{f}} \gets \text{\textit{Set-convex-hull} around } \vec{p}_{\mathrm{f}}, \vec{R}_{\mathrm{f}}$\Comment{Set around end}
		\State $\mathcal{G} \gets \{\}$ \Comment{Initialize graph of convex sets}
		\State $\mathtt{add\_set\_to\_graph}(\mathcal{G}, \mathcal{S}_{\mathrm{0}})$
		\State $\mathtt{connected} \gets \mathtt{add\_set\_to\_graph}(\mathcal{G}, \mathcal{S}_{\mathrm{f}})$
		\State $\mathtt{success} \gets$ False
		\State $\mathtt{refining} \gets$ False
		\State $\pi_{\mathcal{G}, \mathrm{old}} \gets \emptyset$
		\State $i = 1$
		\While{$\mathtt{success} =$ False}
		\If{$\mathtt{connected} =$ True}
		\State $\pi_{\mathcal{G}} \gets \mathtt{dijkstras\_algorithm}(\mathcal{G})$
		\If{$\pi_\mathcal{G} =\pi_{\mathcal{G}, \mathrm{old}}$}
		\State $\mathtt{success} \gets$ True
		\State $\vec{p}_{\mathrm{samples}} \gets \{\}$
		\Else
		\State $\mathtt{refining} \gets $ True
		\State $\vec{p}_{\mathrm{samples}} \gets \{\vec{p}_{a, b} \text{~}|\text{~} \forall (a, b): \mathcal{S}_{a, b} \in \pi_{\mathcal{G}}\}$
		\State $\pi_{\mathcal{G}, \mathrm{old}} = \pi_\mathcal{G}$
		\EndIf
		\Else
		\State $\mathtt{refining} \gets $ False
		\State $\vec{p}_{\mathrm{samples}} \gets \mathrm{sample} \text{~} \vec{p} \in \mathcal{D}_{\mathrm{free}}$
		\EndIf
		\For{$\vec{p} \in \vec{p}_{\mathrm{samples}}$}
		\If{$\mathtt{refining} = $ True}
		\State $\mathcal{S}_{i} \gets \text{\textit{MVIE-fixed-mid} around }\vec{p}$
		\Else
		\State $\mathcal{S}_{i} \gets \text{\textit{MVIE} around }\vec{p}$
		\EndIf
		\State $\mathtt{connected} \gets \mathtt{add\_set\_to\_graph}(\mathcal{G}, \mathcal{S}_{i})$
		\State $i = i + 1$
		\EndFor
		\EndWhile
		\State $\vec{\pi}_{\mathrm{p}}$, $\vec{\pi}_{\mathrm{o}} \gets \mathrm{solve}$~\cref{eq:path_opt} \Comment{Reference paths} \\
		\Return $\vec{\pi}_{\mathrm{p}}$, $\vec{\pi}_{\mathrm{o}}$, $\pi_{\mathcal{G}}$
	\end{algorithmic}
\end{algorithm}

\subsection{Path Planning within the Graph of Convex Sets}%
\label{sub:path_planning}

This section explains the procedure of finding a reference path $\pi$ from the path of sets $\pi_{\mathcal{G}}$ from
\cref{sub:graph_convex_sets}.
The reference path $\pi(\phi)$ consists of
a position reference path $\vec{\pi}_{\mathrm{p}}(\phi)$ and an orientation reference
path $\vec{\pi}_{\mathrm{o}}(\phi)$ in rotation matrix form. Both are assumed
piecewise linear where a linear orientation path is given by a constant spatial angular velocity as in~\cite{oelerichBoundMPCCartesianTrajectory2024}. The path parameter
$\phi$ determines the progress along the path. As mentioned, the set of points
$\vec{p}_{\mathrm{e}, l}(\phi)$, $l = 1, \ldots, L$ define the convex hull of the end-effector geometry. Point $\vec{p}_{\mathrm{e}}(\phi)$ is
the end-effector position for which the position and
orientation paths are planned.
The points $\vec{p}_{\mathrm{e}, l}(\phi)$ are expressed relative to  $\vec{p}_{\mathrm{e}}(\phi)$ as
\begin{equation}
	\vec{p}_{\mathrm{e}, l}(\vec{p}_{\mathrm{e}}(\phi), \vec{R}_{\mathrm{e}}(\phi)) = \vec{p}_{\mathrm{e}}(\phi) +
	\vec{R}_{\mathrm{e}}(\phi) \vec{l}_{\mathrm{e}, l} \text{~,~} l = 1, \ldots, L \text{~,}
\end{equation}
with the constant offset vectors of the
hull points $\vec{l}_{\mathrm{e}, l}$.

The poses that define the piecewise linear paths $\vec{\pi}_{\mathrm{p}}(\phi)$ and
$\vec{\pi}_{\mathrm{o}}(\phi)$ are called via-points. Each intersection set between
two consecutive sets $\mathcal{S}_{\cap, i}$ = $\mathcal{S}_{\mathrm{path}, i}
	\cap \mathcal{S}_{\mathrm{path}, i+1}$, $i = 0, \ldots, N-1$, in $\pi_{\mathcal{G}}$ contains one
via-point. Hence, the piecewise linear position path
\begin{equation}
	\label{eq:position_path}
	\vec{\pi}_{\mathrm{p}}(\phi) = \begin{cases}
		\vec{p}_{0} + (\phi - \phi_{0}) \vec{v}_{0}       & \phi_{0} \leq \phi < \phi_{1}   \\
		% \vec{p}_{1} + (\phi - \phi_{1}) \vec{v}_{1}       & \phi_{1} \leq \phi < \phi_{2}   \\
		\quad\vdots                                       & \quad\vdots                     \\
		\vec{p}_{N-1} + (\phi - \phi_{N-1}) \vec{v}_{N-1} & \phi_{N-1} \leq \phi < \phi_{N} \\
		\vec{p}_{N} = \vec{p}_{\mathrm{f}}                & \phi = \phi_{N}\text{~,}        \\
	\end{cases}
\end{equation}
with the positions $\vec{p}_{i} \in \mathcal{S}_{\cap, i}$ and unit-norm direction vectors
$\vec{v}_{i} = (\vec{p}_{i+1} - \vec{p}_{i})/\lVert\vec{p}_{i+1} -
	\vec{p}_{i}\rVert_{2}$, has exactly one linear
path segment in each convex set $\mathcal{S}_{\mathrm{path}, i}$. Setting $\phi_{i+1} - \phi_{i} = \lVert \vec{p}_{i+1} -\vec{p}_{i}\rVert_{2} $ ensures continuity of the path~\cref{eq:position_path}.
Due to the convexity
property, each segment from $\vec{p}_{i}$ to $\vec{p}_{i+1}$ is entirely contained in
$\mathcal{S}_{\mathrm{path}, i+1}$, and,
thus, the whole piecewise linear position path is contained in
$\pi_{\mathcal{G}}$. Note that this containment is only valid for the end-effector position
$\vec{p}_{\mathrm{e}}(\phi)$
but disregards the extent of the end-effector. In order to account for the end-effector\textquotesingle s
extent, the points $\vec{p}_{\mathrm{e}, l}(\vec{\pi}_{\mathrm{p}}(\phi),
	\vec{\pi}_{\mathrm{o}}(\phi))$ are enforced to be within the
corresponding set. The paths for these points depend on the orientation of the
end-effector. The orientation path $\vec{\pi}_{\mathrm{o}}(\phi)$ starts at
$\vec{\pi}_{\mathrm{o}}(0) = \vec{R}_0$ and ends at
$\vec{\pi}_{\mathrm{o}}(\phi_{\mathrm{f}}) = \vec{R}_{\mathrm{f}}$.
As a simplification, the orientation path is assumed to be
a piecewise linear path with a piecewise constant spatial
angular velocity composed of a constant unit-norm direction
$\vec{\omega}_{\mathrm{ref}}$ and the magnitudes $\alpha_{i}$, i.e.,
\begin{equation}
	\label{eq:orientation_path}
	%\vec{\omega}_{\mathrm{o}}(\phi) = \begin{cases}
	%    \alpha_{0} \vec{\omega}_{\mathrm{ref}} & \vec{\pi}_{\mathrm{p}}(\phi) \in \mathcal{S}_{\mathrm{path},1} \\
	%    \alpha_{1} \vec{\omega}_{\mathrm{ref}} & \vec{\pi}_{\mathrm{p}}(\phi) \in \mathcal{S}_{\mathrm{path},2} \\
	%    \quad\vdots & \quad\vdots \\
	%    \alpha_{N-1} \vec{\omega}_{\mathrm{ref}} & \vec{\pi}_{\mathrm{p}}(\phi) \in \mathcal{S}_{\mathrm{path},N} \\
	\vec{\pi}_{\mathrm{o}}(\phi) = \left\{ \begin{smallmatrix}
		\mathrm{Exp}(\alpha_{0} \frac{\phi - \phi_{0}}{\phi_{1} - \phi_{0}} \vec{\omega}_{\mathrm{ref}})\vec{R}_0          & \phi_{0} \leq \phi < \phi_{1} \\
		% \mathrm{Exp}(\alpha_{1}\frac{\phi - \phi_{1}}{\phi_{2} - \phi_{1}} \vec{\omega}_{\mathrm{ref}})\vec{R}_1           & \phi_{1} \leq \phi < \phi_{2}   \\
		\quad\vdots                                                                                                        & \quad\vdots                   \\
		\mathrm{Exp}(\alpha_{N-1}\frac{\phi - \phi_{N-1}}{\phi_{N} - \phi_{N-1}} \vec{\omega}_{\mathrm{ref}})\vec{R}_{N-1} & \phi_{N-1} \leq \phi < \phi_{N} \\
		\vec{R}_N = \vec{R}_{\mathrm{f}}                                                                                   & \phi = \phi_{N}\text{~.}      \\
	\end{smallmatrix}\right.
\end{equation}
\begin{remark}
	The simplification in~\cref{eq:orientation_path} is introduced to simplify the path-finding problem. This is a performant choice which is applicable to many motions in practice. Future work will explore ways to improve this simplification.
\end{remark}
The direction $\vec{\omega}_{\mathrm{ref}}$ is given
by the shortest connection, i.e., the geodesic between the initial and final
orientation. The intermediate orientations are iteratively defined as $\vec{R}_{i} =
	\mathrm{Exp}(\alpha_{i-1} \vec{\omega}_{\mathrm{ref}})\vec{R}_{i-1}$, $i =
	1, \ldots, N$.
Hence, the path $\pi$ is constructed using~\cref{eq:position_path} and~\cref{eq:orientation_path}
from the via-points $\vec{p}_{i}$ and the magnitudes $\alpha_{i}$ only.
The optimization problem
\begin{equation}
	\label{eq:path_opt}
	\begin{aligned}
		\min_{\vec{p}_{i}, \alpha_{i}} & \sum_{i = 1}^{N-2}  c_{\mathrm{size}, \pi, i} \left(\lVert\vec{p}_{i+1} - \vec{p}_{i}\rVert^{2}_{2} + w_{\alpha}\lVert\alpha_{i+1} - \alpha_{i}\rVert^{2}_{2}\right) \\
		\text{s.t.}\quad               & \vec{p}_{i} \in \mathcal{S}_{\mathrm{\cap}, i}\text{~,~} i = 1, \ldots, N-1                                                                                          \\
		                               & \vec{p}_{\mathrm{e}, l}(\vec{\pi}_{\mathrm{p}}(\phi), \vec{\pi}_{\mathrm{o}}(\phi)) \in \mathcal{S}_{\mathrm{path}, i}                                               \\
		                               & \quad \phi_{i-1} < \phi < \phi_{i} \text{~,~} l = 1, \ldots, L \text{,~} i = 1, \ldots, N                                                                            \\
		                               & \vec{p}_{\mathrm{e}, l}(\vec{\pi}_{\mathrm{p}}(\phi_i),
		\vec{\pi}_{\mathrm{o}}(\phi_i)) \in \mathcal{S}_{\cap, i}                                                                                                                                             \\
		                               & \quad l = 1, \ldots, L \text{,~} i = 1, \ldots, N-1\text{~,}                                                                                                         \\
	\end{aligned}
\end{equation}
with $w_{\alpha} > 0$,
finds these parameters to obtain the shortest reference path $\pi(\phi)$ through $\pi_{\mathcal{G}}$ such that
the end-effector and all hull points remain inside the convex sets $\mathcal{S}_{\mathrm{path}, i}$.
The size cost $c_{\mathrm{size}, \pi, i}$ is computed using~\cref{eq:size_cost}, which favors large sets over small sets.
The first constraint in~\cref{eq:path_opt} ensures that the position path $\vec{\pi}_{\mathrm{p}}(\phi)$ is
contained in the convex sets $\mathcal{S}_{\cap, i}$. The other constraints ensure that the convex hull
of the end-effector is contained
in the convex sets for all values of $\phi$, which needs additional considerations.
With the convex set
$\mathcal{S}_{\mathrm{path}, i}$ for $\phi_{i} < \phi < \phi_{i+1}$, $i = 0, \ldots, N-1$, the hull
points $\vec{p}_{\mathrm{e}, l}(\vec{\pi}_{\mathrm{p}}(\phi),
	\vec{\pi}_{\mathrm{o}}(\phi))$ have distances
\begin{equation}
	\label{eq:dist_half-spaces}
	\vec{d}_{\mathrm{path}, i}(\phi) = \vec{A}_{\mathrm{path}, i} \vec{p}_{\mathrm{e}, l}(\vec{\pi}_{\mathrm{p}}(\phi),
	\vec{\pi}_{\mathrm{o}}(\phi)) - \vec{b}_{\mathrm{path}, i}
\end{equation}
to the half spaces of $\mathcal{S}_{\mathrm{path}, i}$ defined by
$\vec{A}_{\mathrm{path}, i}$ and $\vec{b}_{\mathrm{path}, i}$.
As the orientation path $\vec{\pi}_{\mathrm{o}}(\phi)$ follows the geodesic between
the initial orientation $\vec{R}_{0}$ and desired orientation $\vec{R}_{\mathrm{f}}$, the maximum angle traversed by the
orientation path is \SI{180}{\degree}. Geometrically each hull point moves along a circle in Cartesian space combined with a linear motion due to the linear position path $\vec{\pi}_{\mathrm{p}}$ in~\cref{eq:position_path}. Hence, the function~\cref{eq:dist_half-spaces} is either a convex or a concave function for each element $d_{\mathrm{path}, i, s}$, $s = 1, \ldots, S$, of $\vec{d}_{\mathrm{path}, i}$.
The minima $\vec{\phi}_{\mathrm{min}}^{\mathrm{T}}= [\phi_{\mathrm{min}, 1}, \ldots, \phi_{\mathrm{min}, N}]$ of~\cref{eq:dist_half-spaces} are at
\begin{equation}
	\phi_{\mathrm{min}, i} = \argmin_{\phi_{i-1} < \phi < \phi_i} d_{\mathrm{path}, i, s}(\phi) \text{,~} s = 1, \ldots, S \text{~,}
\end{equation}
which simplifies the second
constraint in~\cref{eq:path_opt} to
\begin{equation}
	\begin{aligned}
		 & \vec{A}_{\mathrm{path}, i} \vec{p}_{\mathrm{e},
		l}(\vec{\pi}_{\mathrm{p}}(\phi_{\mathrm{min}, i}),
		\vec{\pi}_{\mathrm{o}}(\phi_{\mathrm{min}, i})) \leq \vec{b}_{\mathrm{path}, i} \\
		 & \quad l =
		1, \ldots, L \text{~,~} i = 1, \ldots, N.                                       \\
	\end{aligned}
\end{equation}
This results in a finite number of constraints, making~\cref{eq:path_opt} solvable.
Since the gradients of~\cref{eq:position_path} are discontinuous, at the via-points Euler spirals~\cite{talboRailwayTransitionSpiral1915} are used to smooth the path $\vec{\pi}_{\mathrm{p}}(\phi)$.

\section{BoundMPC with Convex Sets as Bounds}%
\label{ssub:mpc_planning}

The BoundMPC~\cite{oelerichBoundMPCCartesianTrajectory2024} framework computes optimal joint-space trajectories in constrained Cartesian spaces along reference paths $\vec{\pi}_{\mathrm{p}}(\phi)$ and $\vec{\pi}_{\mathrm{o}}(\phi)$ using a discrete planning horizon of $M$ time steps, see~\cite{oelerichBoundMPCCartesianTrajectory2024} for a thorough description.
This section describes the adaption of BoundMPC to handle the path $\pi(\phi)$ with the convex sets $\mathcal{S}_{\mathrm{path}, i}$ as path bounds which BoundPlanner computes in~\cref{sec:bound_planner}.

\subsection{Convex Sets as Path Bounds}%
\label{sub:path_bounds}

Having obtained the reference path $\pi(\phi)$ with~\cref{eq:position_path}--\cref{eq:path_opt}, the bounds for the
position path $\vec{\pi}_{\mathrm{p}}(\phi)$ are determined in the following. The original work~\cite{oelerichBoundMPCCartesianTrajectory2024} uses polynomials along basis vectors to limit the deviations from the path.
In contrast, this work uses the constraint
\begin{equation}
	\begin{aligned}
		\vec{g}_{\mathrm{set}} & = \vec{A}_{\mathrm{path}, i} \vec{p}_{\mathrm{tcp}}(t) \leq \vec{b}_{\mathrm{path}, i} \\
		                       & \phi_{i-1} < \phi(t) < \phi_{i} \text{~,~} i = 1, \ldots, N                            \\
	\end{aligned}
	\label{eq:mpc_constraint}
\end{equation}
for bounding deviations in BoundMPC, where $\vec{p}_{\mathrm{tcp}}(t)$ is the position of the end-effector at time $t$. Note that the position $\vec{p}_{\mathrm{tcp}}(t)$ is not necessarily on the path and thus differs from $\vec{p}_{\mathrm{e}}(\phi)$.
% If the position $\vec{p}_{\mathrm{tcp}}(t)$ is exactly on the path $\vec{\pi}_{\mathrm{p}}$ and the orientation of the end-effector is on $\vec{\pi}_{\mathrm{o}}$, then the whole end-effector geometry is collision-free by the design of the reference paths in~\cref{sub:path_planning}. If the current end-effector orientation deviates from the path, additional considerations are necessary.

% One option is the specification of orientation bounds for the orientation path $\vec{\pi}_{\mathrm{o}}(\phi)$. Such bounds may be specified as bounds on the RPY angles in the BoundMPC framework. However, using this kind of bounds for obstacle avoidance is challenging because they depend on the end-effector position, i.e., for different positions within the convex set, different orientation bounds are needed. Furthermore, the SO(3) orientation space is a manifold with nonzero curvature~\cite{cohnNonEuclideanMotionPlanning2023} meaning that convexity is not preserved for sets that are transferred between SO(3) and the task space. Hence, determining such bounds is computationally expensive.
The planned reference path $\pi(\phi)$ ensures a collision-free robot motion only for the end-effector along the reference position path $\vec{\pi}_{\mathrm{p}}(\phi)$ with the reference orientation path $\vec{\pi}_{\mathrm{o}}(\phi)$. In order to consider the entire robot kinematics for general robot manipulators, this work uses a novel formulation for collision avoidance based on convex sets. This is presented here for $R$ arbitrary points on the robot $\vec{p}_{\mathrm{c}, r}(t)$, $r = 1, \ldots, R$, that need to be collision checked at all times $t$.
The collision constraint for the trajectory planning at time $t$ becomes simply
\begin{equation}
	\vec{A}_{\mathrm{c}, r} \vec{p}_{\mathrm{c}, r} \leq \vec{b}_{\mathrm{c}, r} \text{~,~} r = 1, \ldots, R \text{~,}
	\label{eq:set_collision_avoidance}
\end{equation}
with individual convex sets $\mathcal{S}_{\mathrm{c}, r}$ defining the half spaces $\vec{A}_{\mathrm{c}, r}$ and $\vec{b}_{\mathrm{c}, r}$.
The sets $\mathcal{S}_{\mathrm{c}, r}$ are found by solving the \textit{Set-convex-hull} problem.
The points $\vec{p}_{\mathrm{ch}, j}$ defining the convex hull are set to the position $\vec{p}_{\mathrm{ch}, 1} = \vec{p}_{\mathrm{c}, r}(t)$ at the current time $t$ and the estimated position $\vec{p}_{\mathrm{ch}, 2} = \hat{\vec{p}}_{\mathrm{c}, r}(t_{\mathrm{h}})$ at time $t_{\mathrm{h}}$ at the end of the planning horizon of the previous MPC iteration, respectively. Note that the convex hull of two points is a line which makes~\cref{eq:proj_conv_hull} analytically solvable.
Consequently, the formulation~\cref{eq:set_collision_avoidance} defines the collision-free space for each point $\vec{p}_{\mathrm{c}, r}$ instead of defining the obstacle-occluded space as is commonly done in robotics~\cite{oelerichBoundMPCCartesianTrajectory2024, beckModelPredictiveTrajectory2024, vuFastMotionPlanning2020}. Convex sets are easy to formulate mathematically and to compute.

The constraint formulation~\cref{eq:set_collision_avoidance} is independent of the number of obstacles $\mathcal{O}_{k}$ as it only requires the convex sets $\mathcal{S}_{\mathrm{c}, r}$. While this approach is generally more conservative than describing the obstacle-occluded space, it yields good solutions in practice. The computation of the sets depends on the number of obstacles and is generally fast, even for a large number of obstacles~\cite{wangFastIterativeRegion2024, spahnCoupledMobileManipulation2021}.
\review{By considering the currently predicted position $\hat{\vec{p}}_{\mathrm{c}, r}(t_{\mathrm{h}})$, it considers the current motion of the robot contrary to the formulation in~\cite{spahnCoupledMobileManipulation2021}.}
% This becomes especially relevant when working with point clouds since they consist of hundreds or thousands of points.

\begin{remark}
	The points $\vec{p}_{\mathrm{c}, r}$ may be used to describe the robot collision model using spheres. As it is commonly done, the size of the obstacles is extended by the radius of the sphere corresponding to the robot dimensions at $\vec{p}_{\mathrm{c}, r}$.
\end{remark}

\subsection{Horizon splitting}%
\label{ssub:horizon_splitting}

The path bounds in~\cite{oelerichBoundMPCCartesianTrajectory2024} are described by polynomials along basis directions. In contrast, this work uses the constraint~\cref{eq:mpc_constraint}.
However, directly enforcing~\cref{eq:mpc_constraint} is undesirable since the constraint is not continuous. Instead, the switching points between two consecutive path sets are determined before the optimization such that the BoundMPC optimization remains continuous. This is implemented based on the idea of horizon splitting~\cite{beckModelPredictiveTrajectory2024}. Imagine that the current horizon of the end-effector position trajectory $\vec{p}_{\mathrm{horizon}, m}$, $m = 1, \ldots, M$, spans two path sets $\mathcal{S}_{\mathrm{path}, i}$ and $\mathcal{S}_{\mathrm{path}, i+1}$. The splitting index
\begin{equation}
	\begin{aligned}
		m_{\mathrm{s}} = \min_{m} & (\phi > \phi_{i} - \epsilon_{\phi}  \land \vec{p}_{\mathrm{horizon}, m} \in \mathcal{S}_{\mathrm{path}, i} \\
		                          & \land \vec{p}_{\mathrm{horizon}, m} \in \mathcal{S}_{\mathrm{path}, i+1})
	\end{aligned}
	\label{eq:split_idx}
\end{equation}
with tolerance $\epsilon_{\phi} \geq 0$ is determined by the set membership of the end-effector position at each time step in the horizon.
This leads to the constraint
\begin{equation}
	\vec{g}_{\mathrm{s}} = \left\{\begin{matrix}
		\vec{A}_{\mathrm{path}, i} \vec{p}_{\mathrm{horizon}, m} \leq \vec{b}_{\mathrm{path}, i} \text{,~} m < m_{\mathrm{s}} \\
		\vec{A}_{\mathrm{path}, i+1} \vec{p}_{\mathrm{horizon}, m} \leq \vec{b}_{\mathrm{path}, i+1} \text{,~} m \geq m_{\mathrm{s}} \text{.}
	\end{matrix}\right.
	\label{eq:split_constraint}
\end{equation}
Furthermore, a terminal constraint is added based on the first set that is outside the horizon $\mathcal{S}_{\mathrm{path, i+2}}$ which guides the robot along the path. Particularly, the terminal constraint limits the path deviations such that the robot can transition into $\mathcal{S}_{\mathrm{path}, i+2}$, guiding it along the path.
Lastly, constraint~\cref{eq:set_collision_avoidance} is added to account for collisions of the kinematic chain.

\subsection{Replanning}%
\label{sub:replanning}

% \begin{figure}[t]
% 	\centering
% 	\def\svgwidth{0.55\linewidth}
% 	\import{inkscape}{replanning.pdf_tex}
% 	\caption{Schematic drawing of the replanning procedure where the replanning happens in $\mathcal{S}_{\mathrm{cur}}$. The replanning set $\mathcal{S}_{\mathrm{replan}}$ is computed to include $\vec{p}_{\mathrm{horizon}, 1}$ and $\vec{p}_{\mathrm{horizon, max}}$.}
% 	\label{fig:replanning}
% \end{figure}

The fast computation of the reference paths $\vec{\pi}_{\mathrm{p}}(\phi)$ and $\vec{\pi}_{\mathrm{o}}(\phi)$ combined with the replanning capabilities of BoundMPC~\cite{oelerichBoundMPCCartesianTrajectory2024} enables online replanning of the reference paths. This procedure is illustrated in~\cref{fig:planner_scheme}.
To ensure feasibility after the replanning event, it is important to account for the current robot state. The current planning horizon computed by BoundMPC includes $M$ discretized end-effector positions $\vec{p}_{\mathrm{horizon}, m}$, $m = 1, \ldots, M$. The current end-effector position $\vec{p}_{\mathrm{horizon}, 1}$ is contained in the current set $\mathcal{S}_{\mathrm{cur}}$ defined by $\vec{A}_{\mathrm{cur}}$ and $\vec{b}_{\mathrm{cur}}$. Then, the point
\begin{equation}
	\begin{aligned}
		\vec{p}_{\mathrm{horizon, max}} = & \max_{m}\vec{p}_{\mathrm{horizon}, m}                                                              \\
		                                  & \text{ s.t.} \quad \vec{A}_{\mathrm{cur}}\vec{p}_{\mathrm{horizon}, m} \leq \vec{b}_{\mathrm{cur}}
	\end{aligned}
	\label{eq:replan_point}
\end{equation}
is the temporally farthest-away point in the planning horizon that is still within the current set $\mathcal{S}_{\mathrm{cur}}$.

The starting point during the replanning procedure is set to $\vec{p}_{0} = \vec{p}_{\mathrm{horizon}, 1}$, for which a starting set $\mathcal{S}_{\mathrm{replan}}$ is computed using the \textit{Set-convex-hull} approach with $\vec{p}_{\mathrm{l, 0}} = \vec{p}_{\mathrm{horizon}, 1}$ and $\vec{p}_{\mathrm{l, 1}} = \vec{p}_{\mathrm{horizon, max}}$ such that the first and last point in the horizon are inside the set $\mathcal{S}_{\mathrm{replan}}$. This set is then used as $\mathcal{S}_0$ in~\cref{alg:bound_planner}. After replanning, the path $\vec{\pi}_{\mathrm{p}}(\phi)$ should be extended at $\vec{p}_{\mathrm{0}}$ to ensure proper projection onto the new reference path with BoundMPC. This is indicated visually in~\cref{fig:planner_scheme}.

\section{Experiments}\label{sec:experiments} % (fold)

The proposed planning method is evaluated in two scenarios.
The first scenario in~\cref{sub:scen_box} requires the 7-DoF KUKA LBR iiwa 14 R820 robot to turn around and reach into a box. In the second scenario in~\cref{sub:scen_replan}, the manipulator must grasp an object from a shelf but the object changes its position during the approaching motion such that online replanning is required.

\subsection{Scenario 1: Reach into box}%
\begin{figure}
	\centering
	\def\axisdefaultwidth{\linewidth}
	\def\axisdefaultheight{0.5\linewidth}
	\input{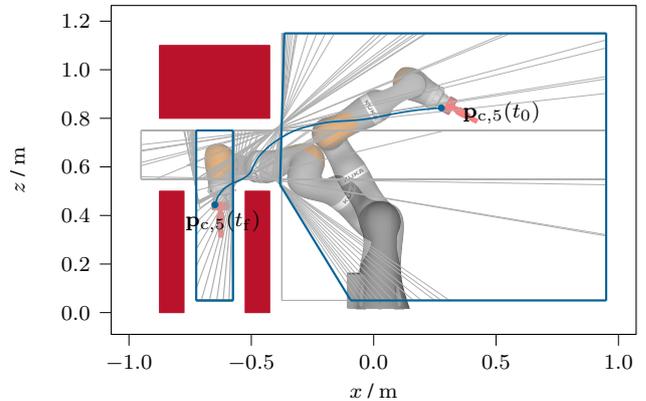}
	\caption{Open-box scenario: Collision avoidance sets for the point $\vec{p}_{\mathrm{c}, 5}$ on the robot\textquotesingle s kinematic chain. The set $\mathcal{S}_{\mathrm{c}, 5}$ for the initial time $t_{0}$ and the final time $t_{\mathrm{f}}$ is shown in green and at intermediate time steps in gray. The gray lines illustrate the evolution of $\mathcal{S}_{\mathrm{c}, 5}$ over time with increasing line width.} \label{fig:scen_open_box_ca}
\end{figure}

\label{sub:scen_box}
The robot must reach into a box where an additional obstacle above the box significantly complicates the trajectory planning as depicted in~\cref{fig:scen_open_box}. It has to move from an open space into a restricted space in the box.
The depicted path sets $\mathcal{S}_{\mathrm{path}, i}$, $i = 0, 1, 2$, cover a large portion of the free space $\mathcal{D}_{\mathrm{free}}$. Using BoundPlanner and BoundMPC, the robot\textquotesingle s end-effector trajectory $\vec{p}_{\mathrm{tcp}}(t)$ successfully traverses the convex sets to the final position $\vec{p}_{\mathrm{f}}$ with the final orientation $\vec{R}_{\mathrm{f}}$. Large deviations of the trajectory $\vec{p}_{\mathrm{tcp}}(t)$ from the reference path $\vec{\pi}_{\mathrm{p}}(\phi)$ are observed to account for the kinematic constraints of the robot.
Collision avoidance is considered for the end-effector by constraining the position $\vec{p}_{\mathrm{tcp}}(t)$ to the path sets $\mathcal{S}_{\mathrm{path}, i}$ and for the rest of the kinematic chain using the constraint~\cref{eq:set_collision_avoidance} with $R = 5$ points $\vec{p}_{\mathrm{c}, r}$ along the chain. The points $\vec{p}_{\mathrm{c}, r}$ are chosen on the joint axes 3, 4, 5, 6, and between joint 6 and the end-effector. The evolution of the collision-free set $\mathcal{S}_{\mathrm{c}, 5}$ for the point $\vec{p}_{\mathrm{c}, 5}$ over time is depicted in~\cref{fig:scen_open_box_ca}. The margins around the obstacles account for the size of the robot around $\vec{p}_{\mathrm{c}, 5}$. The convex sets capture the space around the point and allow the robot to move into the strongly constrained space around the final pose, showing the effectiveness of the constraint~\cref{eq:set_collision_avoidance} and the iterative update of $\mathcal{S}_{\mathrm{c}, 5}$. Due to their convexity, the sets $\mathcal{S}_{\mathrm{c}, 5}$ are a conservative approximation of the free space around $\vec{p}_{\mathrm{c}, 5}$, but this does not degrade performance in practice.

BoundPlanner with BoundMPC is also compared to state-of-the-art approaches, namely the graph-of-convex-sets (GCS) approach from~\cite{marcucciMotionPlanningObstacles2023}, VP-STO~\cite{jankowskiVPSTOPointbasedStochastic2023}, and a bidirectional RRT~\cite{lavallePlanningAlgorithms2006} approach with a tracking MPC that minimizes the tracking error with respect to the reference trajectory.
The GCS method requires the final joint configuration as a goal for trajectory planning and seed points in joint space to explore the collision-free space. Both inputs are generally hard to provide and are not required by BoundPlanner. Hence, for the comparison, the seed points are provided manually as in~\cite{marcucciMotionPlanningObstacles2023} and consist of six configurations, including the start and end configuration.
VP-STO~\cite{jankowskiVPSTOPointbasedStochastic2023} is a global planner that directly plans joint trajectories based on a Cartesian description of the obstacles. Collision freedom is enforced in the cost function. The final joint configuration is provided to VP-STO since it does not converge otherwise.
The RRT approach uses a bidirectional RRT~\cite{lavallePlanningAlgorithms2006} to plan a collision-free reference path in position space. The corresponding orientations are computed as interpolations between the start and end orientation, similar to BoundPlanner. The reference trajectory, consisting of position and orientation, is obtained as a spline~\cite{toussaintSequenceConstraintsMPCReactive2022}. A trajectory-tracking MPC is then used to track the reference trajectory to minimize the tracking error. Collisions are avoided using potential functions~\cite{beckModelPredictiveTrajectory2024}.

The computed trajectories for all methods are compared in~\cref{fig:scen_open_box}. BoundPlanner with BoundMPC and VP-STO create short Cartesian trajectories and reach $\vec{p}_{\mathrm{f}}$. The GCS method also converges to $\vec{p}_{\mathrm{f}}$ but has a longer trajectory in Cartesian space. Finally, the RRT+MPC method gets stuck in a local minimum in front of the box and cannot reach the final pose.
In order to further evaluate the robustness of the methods, eight different placements of the open box are considered and evaluated in~\cref{tab:scen_open_box_comp} for all methods. The lengths $l_{\mathrm{traj, p}}$ and $l_{\mathrm{traj, o}}$ for the Cartesian end-effector trajectory indicate short trajectories for our method due to the Cartesian planning. The planning time $t_{\mathrm{plan}}$ is the lowest for our method enabling replanning of the movement online. Furthermore, no failures occur using our method, whereas RRT+MPC often gets stuck in local minima due to the simple tracking objective. BoundPlanner only specifies a planning volume for BoundMPC instead of minimizing the tracking error, allowing more freedom during the robot\textquotesingle s motion.
Hence, our method reliably computes collision-free paths and tracking volumes for the entire kinematic chain in constrained environments.
The GCS planner fails once despite providing the desired end configuration in joint space and the six seed configurations because the convex sets do not allow for a connection between the start and end configuration. Also, GCS generally has longer Cartesian trajectories with long planning times $t_{\mathrm{plan}}$.
VP-STO solves the task with shorter trajectory durations $T_{\mathrm{traj}}$  but requires longer planning times $t_{\mathrm{plan}}$, and is therefore unsuitable for online replanning.

\begin{table}
	\addtolength\abovecaptionskip{-5pt}
	\caption{Open-box scenario: Comparison of our method with GCS, RRT+MPC and VP-STO for eight different box placements. The reported values are the mean values of the end-effector trajectory length for position $l_{\mathrm{traj, p}}$ and orientation $l_{\mathrm{traj, o}}$, the planning time $t_{\mathrm{plan}}$, and the trajectory duration $T_{\mathrm{traj}}$.}\label{tab:scen_open_box_comp}
	\centering
	\begin{tabular}[c]{ccccc}
		\hline
		                                      & \multicolumn{1}{c}{\textbf{Ours}} & \multicolumn{1}{c}{\textbf{GCS}} & \multicolumn{1}{c}{\textbf{RRT + MPC}} & \multicolumn{1}{c}{\textbf{VP-STO}} \\
		\hline
		$l_{\mathrm{traj, p}}$ / \si{\meter}  & 1.07                              & 3.56                             & \textbf{0.89}                          & 1.45                                \\
		$l_{\mathrm{traj, o}}$ / \si{\degree} & 140                               & 659                              & \textbf{130}                           & 188                                 \\
		$t_{\mathrm{plan}}$ / \si{\second}    & \textbf{0.11}                     & 233.15                           & 0.51                                   & 9.3                                 \\
		$T_{\mathrm{traj}}$ / \si{\second}    & 5.3                               & 13.67                            & 5.4                                    & \textbf{4.21}                       \\
		Failures                              & \textbf{0}                        & 1                                & 4                                      & \textbf{0}                          \\
		\hline
	\end{tabular}
\end{table}

\begin{figure}
	\centering
	\addtolength\abovecaptionskip{-15pt}
	\def\axisdefaultwidth{\linewidth}
	\def\axisdefaultheight{0.7\linewidth}
	\input{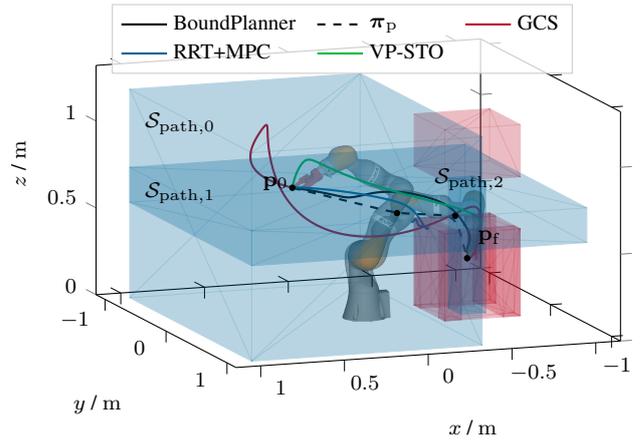}
	\caption{Open-box scenario: The robot has to turn around and reach in the box with its end-effector. The convex obstacles $\mathcal{O}_{k}$, $k = 1, \ldots, 5$, are depicted in red and the convex planning sets $\mathcal{S}_{\mathrm{path}, i}$, $i = 0, \ldots, 2$, in blue. The end-effector trajectories $\vec{p}_{\mathrm{tcp}}(t)$ of different methods as well as the reference path $\vec{\pi}_{\mathrm{p}}(\phi)$ of BoundPlanner are shown.}
	\label{fig:scen_open_box}
\end{figure}

% \begin{figure}
% 	\centering
% 	\def\axisdefaultwidth{\linewidth}
% 	\def\axisdefaultheight{0.7\linewidth}
% 	\input{tikz/scen_reach.tex}
% 	\caption{Scenario}
% 	\label{fig:scen_reach}
% \end{figure}

% \begin{figure}
% 	\centering
% 	\def\axisdefaultwidth{\linewidth}
% 	\def\axisdefaultheight{0.7\linewidth}
% 	\input{tikz/scen_open_box_comp.tex}
% 	\caption{Open-box scenario: Comparison to other methods. }
% 	\label{fig:scen_open_box_comp}
% \end{figure}

\begin{table}
	\addtolength\abovecaptionskip{-5pt}
	\caption{Replanning scenario: Comparison of BoundPlanner with RRT+MPC for 20 random replanning events.}\label{tab:scen_replan_comp}
	\centering
	\begin{tabular}[c]{ccccccc}
		\hline
		                                   & \multicolumn{3}{c}{\textbf{BoundPlanner}} & \multicolumn{3}{c}{\textbf{RRT+MPC}}                             \\
		                                   & min                                       & avg                                  & max  & min  & avg  & max  \\
		\hline
		$t_{\mathrm{plan}}$ / \si{\second} & 0.04                                      & 0.1                                  & 0.17 & 0.02 & 0.54 & 2.06 \\
		Number of collisions               & \multicolumn{3}{c}{0}                     & \multicolumn{3}{c}{3}                                            \\
		\hline
	\end{tabular}
\end{table}

\subsection{Scenario 2: Grasp from shelf with replanning}%
\label{sub:scen_replan}

\begin{figure}
	\addtolength\abovecaptionskip{-15pt}
	\centering
	\def\axisdefaultwidth{\linewidth}
	\def\axisdefaultheight{0.7\linewidth}
	\input{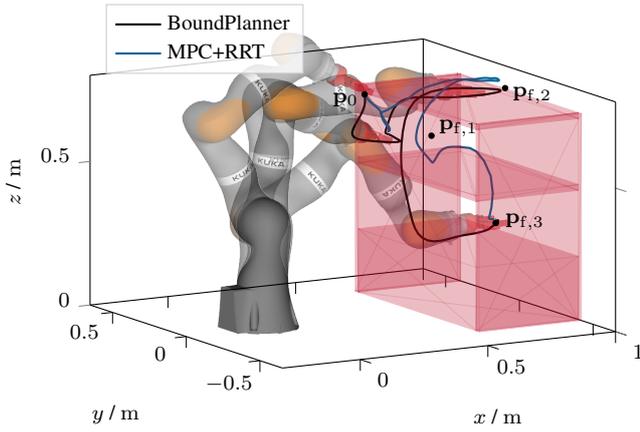}
	\caption{Replanning scenario: The robot has to grasp an object from a shelf, but the object changes its position from $\vec{p}_{\mathrm{f}, 1}$ to $\vec{p}_{\mathrm{f}, 2}$ and finally to $\vec{p}_{\mathrm{f}, 3}$ during the robot\textquotesingle s motion. The convex obstacles $\mathcal{O}_{k}$, $k = 1, \ldots, 5$, are depicted in red. The end-effector trajectories $\vec{p}_{\mathrm{tcp}}(t)$ of BoundPlanner and MPC+RRT are shown.}
	\label{fig:scen_replan}
\end{figure}

In the scenario visualized in~\cref{fig:scen_replan}, the robot grasps an object in different locations that change during the approach motion and require replanning. BoundPlanner with BoundMPC and RRT+MPC are compared, as GCS and VP-STO exhibit too long planning times $t_{\mathrm{plan}}$.
The end-effector trajectories in~\cref{fig:scen_replan} for both methods contain two exemplary replanning events. At the start, the robot plans toward $\vec{p}_{\mathrm{f}, 1}$, and shortly before arriving, the final point is changed to $\vec{p}_{\mathrm{f}, 2}$ that is repeated for the final replanning toward $\vec{p}_{\mathrm{f}, 3}$. Both methods perform the replanning quickly and reach the final point $\vec{p}_{\mathrm{f}, 3}$. However, the RRT+MPC method plans a trajectory that collides with the shelf after the second replanning. This collision is attributed to the robot prioritizing the minimization of the tracking error over the collision avoidance. This is avoided by BoundMPC as the convex sets used for collision avoidance do not allow for such trade-offs resulting in safe and collision-free trajectories for the robot.

The robustness of random replanning events is further evaluated for 20 replannings, performed for random positions. The replannings happen shortly before reaching the desired pose. The results in~\cref{tab:scen_replan_comp} show that the replanning times $t_{\mathrm{plan}}$ of BoundPlanner are generally faster, and the trajectories are collision-free. The collisions of MPC+RRT are attributed to the same phenomenon as in~\cref{fig:scen_replan}.

\section{Conclusions}
\label{sec:conclusions}

A novel Cartesian path planner based on convex sets, called BoundPlanner, is proposed in this work for fast planning in cluttered environments for robot manipulators. The online trajectory planner BoundMPC was extended to handle convex sets as bounds, allowing a robot to safely execute complex manipulation tasks in confined spaces including collision avoidance for the entire robot kinematics. The proposed approach was compared with state-of-the-art methods on a 7-DoF robot manipulator in two scenarios. The first scenario requires the robot to grasp into a box in a strongly constrained environment where the collision avoidance is demonstrated and the fast planning times compared to global planners are highlighted.
The second scenario demonstrates the replanning capabilities of BoundPlanner to create safe and performant trajectories online.
Future work will embed BoundPlanner in a higher-level decision-making framework to solve long-horizon manipulation tasks reliably.
\review{There is no guarantee that a joint trajectory exists along the bounded path, which needs to be investigated further.}

%%%%%%%%%%%%%%%%%%%%%%%%%%%%%%%%%%%%%%%%%%%%%%%%%%%%%%%%%%%%%%%%%%%%%%%%%%%%%%%%

%%%%%%%%%%%%%%%%%%%%%%%%%%%%%%%%%%%%%%%%%%%%%%%%%%%%%%%%%%%%%%%%%%%%%%%%%%%%%%%%

%%%%%%%%%%%%%%%%%%%%%%%%%%%%%%%%%%%%%%%%%%%%%%%%%%%%%%%%%%%%%%%%%%%%%%%%%%%%%%%%

\bibliographystyle{IEEEtran}
\bibliography{IEEEabrv,bound_planner_fix}  % .bib

\end{document}